\documentclass[11pt,a4paper]{article}
\usepackage[hyperref]{emnlp-ijcnlp-2019}
\usepackage{graphicx}
\usepackage{times}
\usepackage{latexsym}
\usepackage{gb4e}
\usepackage{balance}
\noautomath

\usepackage{url}

\aclfinalcopy 

\newcommand{\numLangs}{26\ }

\title{Does {BERT} agree? Evaluating knowledge of\\ structure dependence through agreement relations}

\author{Geoff Bacon \\
  Department of Linguistics \\
  University of California, Berkeley \\
  {\tt bacon@berkeley.edu} \\\And
  Terry Regier \\
  Department of Linguistics and \\
  Cognitive Science Program \\
  University of California, Berkeley \\
  {\tt terry.regier@berkeley.edu} \\}

\date{}

\begin{document}
\maketitle
\begin{abstract}
  Learning representations that accurately model semantics is an important goal of natural language processing research. Many semantic phenomena depend on syntactic structure. Recent work examines the extent to which state-of-the-art models for pre-training representations, such as BERT, capture such structure-dependent phenomena, but is largely restricted to one phenomenon in English: number agreement between subjects and verbs. We evaluate BERT's sensitivity to four types of structure-dependent agreement relations in a new semi-automatically curated dataset across \numLangs languages. We show that both the single-language and multilingual BERT models capture syntax-sensitive agreement patterns well in general, but we also highlight the specific linguistic contexts in which their performance degrades.
\end{abstract}

\section{Introduction}

Learning general-purpose sentence representations which accurately model sentential semantic content is a current goal of natural language processing research \cite{subramanian2018learning, conneau-EtAl:2017:EMNLP2017,towards,kiros2015skip}. A prominent and successful approach is to pre-train neural networks to encode sentences into fixed length vectors \cite{P18-1198,DBLP:journals/corr/abs-1710-04334}, with common architecture choices based on recurrent neural networks \cite{elman1990finding,hochreiter1997long}, convolutional neural networks, or transformers \cite{vaswani2017attention}. Many core linguistic phenomena that one would like to model in general-purpose sentence representations depend on syntactic structure \cite{chomsky1969aspects,everaert2015structures}. Despite the fact that none of the aforementioned architectures have explicit syntactic structural representations, there is some evidence that these models can approximate such structure-dependent phenomena under certain conditions \cite{N18-1108,mccoy2018revisiting,linzen2016assessing,bowman2015tree}, in addition to their widespread success in practical tasks. 

The recently introduced BERT model \cite{devlin2018bert}, which is based on transformers, achieves state-of-the-art results on eleven natural language processing tasks. In this work, we assess BERT's ability to learn structure-dependent linguistic phenomena of agreement relations. To test whether BERT is sensitive to agreement relations, we use the cloze test \cite[also called the ``masked language model'' objective]{taylor1953cloze}, in which we mask out one of two words in an agreement relation and ask BERT to predict the masked word, one of the two tasks on which BERT is initially trained.

\citet{goldberg2019assessing} adapted the experimental setup of \citet{linzen2016assessing}, \citet{N18-1108} and \citet{marvin-linzen-2018-targeted} to use the cloze test to assess BERT's sensitivity to number agreement in English subject-verb agreement relations. The results showed that the single-language BERT model performed surprisingly well at this task (above 80\% accuracy in all experiments), even when there were multiple ``distractors'' in the sentence (other nouns that differed from the subject in number). This suggests that BERT is actually learning to approximate structure-dependent computation, and not simply relying on flawed heuristics.

However, English subject-verb agreement is a rather restricted phenomenon, with the majority of verbs having only two inflected forms and only one morphosyntactic feature (number) involved. To what extent does Goldberg's \citeyearpar{goldberg2019assessing} result hold for subject-verb agreement in other languages, including more morphologically rich ones, as well as for other types of agreement relations? Building on Goldberg's \citeyearpar{goldberg2019assessing} work, we expand the experiment to \numLangs languages and four types of agreement relations, which include more challenging examples.


In Section 2, we define what is meant by agreement relations and outline the particular agreement relations under study. Section 3 introduces our newly curated cross-linguistic dataset of agreement relations, while section 4 discusses our experimental setup. We report the results of our experiments in section 5. All data and code are available at \url{https://github.com/geoffbacon/does-bert-agree}.

\section{Structure-dependent agreement relations} \label{section:agreement}

Agreement phenomena are an important and cross-linguistically common property of natural languages, and as such have been extensively studied in syntax and morphology \cite{corbett2006agreement}.\footnote{For a comprehensive bibliography, see \url{http://www.smg.surrey.ac.uk/projects/agreement/bibliography/}.} Languages often express grammatical features, such as number and gender, through inflectional morphology. An agreement relation is a morphophonologically overt co-variance in feature values between two words in a syntactic relationship \cite{Preminger:2014}. In other words, agreement refers to when the morphosyntactic features of one word are reflected in its syntactic dependents. In this way, agreement relations are overt markers of covert syntactic structure. Thus, evaluating a model's ability to capture agreement relations is also an evaluation of its ability to capture syntactic structure.

Following \citet{corbett2003agreement}, we call the syntactically dependent word the ``target'' of the agreement relation, and the word with which it agrees we call the ``controller''. An example of an agreement relation in English is given in (\ref{example:english}), in which the inflected form of the verb \textsc{be} (\emph{are}) reflects the plural number of its syntactic head \emph{keys}. In all examples in this section, the controller and target are given in bold. In this example, \emph{keys} is the controller and \emph{are} is the target of the agreement relation. 

\begin{exe}
\ex The \textbf{keys} to the door \textbf{are} on the table.
\label{example:english}
\end{exe}

The agreement relation in (\ref{example:english}) is between a subject and its verb, but there are other types of agreement relations. In addition to subject-verb agreement, three other types of agreement relations are cross-linguistically common: agreement of noun with i) determiner, ii) attributive adjective and iii) predicate adjective \cite{baker_2008}. The latter two types are distinguished by whether the adjective modifies the noun within a noun phrase or whether it is predicated of the subject of a clause. The first two types are sometimes categorized as nominal concord rather than agreement, but for our purposes this is merely a difference in terminology.

The morphosyntactic feature in the agreement relation in (\ref{example:english}) is number, a feature that is cross-linguistically common in agreement systems. In addition to number, the most commonly involved in agreement relations are gender, case and person \cite{baker_2008}.

With its comparatively limited inflectional morphology, English only exhibits subject-verb and determiner agreement (in demonstratives, ``this'' vs. ``these'') and even then only agrees for number. Languages with richer inflectional morphology tend to display more agreement types and involve more features. French, for example, employs all four types of agreement relations. Examples are given in (\ref{example:verb})-(\ref{example:predicated}). The subject and verb in (\ref{example:verb}) agree for number, while the noun and determiner in (\ref{example:determiner}), the noun and attributive adjective in (\ref{example:modifying}) and the subject and predicated adjective in (\ref{example:predicated}) agree for both number and gender.


\begin{exe}
\ex 
\gll Les \textbf{cl\'es} de la porte se \textbf{trouvent} sur la table.\\ \\
\trans `The keys to the door are on the table.'
\label{example:verb}
\end{exe}

\begin{exe}
\ex 
\gll Je peux voir \textbf{les} \textbf{cl\'es}.\\ \\ 
\trans `I can see the keys.'
\label{example:determiner}
\end{exe}

\begin{exe}
\ex 
\gll Je ne veux plus les \textbf{cl\'es} totalement \textbf{cass\'ees}.\\ \\
\trans `I no longer want the completely broken keys.'
\label{example:modifying}
\end{exe}

\begin{exe}
\ex 
\gll Les \textbf{cl\'es} de la porte sont \textbf{cass\'ees}.\\ \\ 
\trans `The keys to the door are broken.'
\label{example:predicated}
\end{exe}


Previous work using agreement relations to assess knowledge of syntactic structure in modern neural networks has focussed on subject-verb agreement in number \cite{goldberg2019assessing,N18-1108,linzen2016assessing}. In our work, we study all four types of agreement relations and all four features discussed above. Moreover, previous work using any method to assess BERT's knowledge of syntactic structure has focussed exclusively on the single-language English model \cite{hewitt2019structural,goldberg2019assessing,tenney2019bert,lin2019open,jawahar2019does,clark2019does}. We expand this line of work to \numLangs languages. Not all languages in our sample exhibit all four types of agreement nor use all four features examined, but they all exhibit at least one of the agreement types involving at least one of the features.

\section{Data}

Our study requires two types of data. First, we need sentences containing agreement relations. We mask out one of the words in the agreement relation and ask BERT to predict the masked word. We are interested in BERT's ability to predict words that respect the agreement relation, that is, words which share the morphosyntactic features of the word with which it agrees. To measure this, we need to know the feature values for each word in BERT's vocabulary. This is our second type of data. Throughout this paper, we refer to the first type of data as the cloze data, and the second as the feature data.

In the design of our datasets, we followed two principles. First, we chose data sources that are available across multiple languages, because we are interested in cross-linguistic generality. The languages in this study are those with sufficiently large data sources that also appear in the multilingual BERT model. Second, we use naturally-occurring data (cf. \citet{marvin-linzen-2018-targeted}).

\subsection{Cloze data}

We sourced our cloze data from version 2.4 of the Universal Dependencies treebanks \cite[UD]{nivre2016universal}. The UD treebanks use a consistent schema across all languages to annotate naturally occurring sentences at the word level with rich grammatical information. We used the part-of-speech and dependency information to identify potential agreement relations. Specifically, we identified all instances of subject-verb, noun-determiner, noun-attributive adjective and subject-predicate adjective word pairs. We then used the morphosyntactic annotations for number, gender, case and person to filter out word pairs that disagree due to errors in the underlying data source (e.g.\ one is annotated as plural while the other is singular) or that are not annotated for any of the four features. 


This method is language-agnostic, but due to errors in the underlying UD corpora, yielded some false positives (e.g.\ predicate adjective agreement in English). To correct for this, we consulted reference grammars of each language to note which of the four types of agreement exist in the language. We removed all examples that are of the wrong type for the language (8\% of harvested examples). Across the \numLangs languages, we curated almost one million cloze examples. Their breakdown across agreement type and language is shown in Tables 1 and 2.

\begin{table}[h]
\centering
\begin{tabular}{lc}
  Agreement type & \# cloze\\
  \hline
  attributive adjective & 351,300 \\
  determiner & 349,073 \\
  subject-verb & 253,820 \\
  predicate adjective & 26,696 \\ \hline
  total & 980,889
\end{tabular}
\caption{Number of cloze examples per agreement type in our new cross-linguistic dataset on agreement relations. Previous work has largely focused on subject-verb agreement in English.}
\end{table}

In all four types of agreement studied, the controller of the agreement is a noun or pronoun, while the target can be a determiner, adjective or verb. Because of this part-of-speech restriction, we chose to mask out the controller in every cloze example so that BERT is evaluated against the same vocabulary across all four types. This also means that we only need to collect feature data on nouns and pronouns.


\begin{table}[!ht]
\centering
\begin{tabular}{lcc}
  Language & \# cloze & \# feature bundles\\
  \hline
 Russian & 144,458 & 2,404 \\
 Italian & 137,268 & 2,479 \\
 French & 117,769 & 3,384 \\
 Catalan & 88,053 & 1,753 \\
 English & 60,060 & 6,743 \\
 Dutch & 52,569 & 1,531 \\
 Latin & 50,053 & 1,044 \\
 Polish & 47,513 & 2,011 \\
 Portuguese & 47,038 & 2,107 \\
 Finnish & 36,705 & 1,167 \\
 Romanian & 29,746 & 1,330 \\
 Norwegian & 29,059 & 1,393 \\
 Hindi & 22,959 & 402 \\
 Croatian & 21,835 & 1,141 \\
 Persian & 14,238 & 985 \\
 Greek & 14,017 & 216 \\
 Ukrainian & 13,929 & 1,206 \\
 Swedish & 10,889 & 1,611 \\
 Hebrew & 10,809 & 338 \\
 Danish & 9,432 & 1,330 \\
 Urdu & 8,139 & 547 \\
 Basque & 4,132 & 267 \\
 Turkish & 3,155 & 846 \\
 Irish & 3,030 & 259 \\
 Afrikaans & 2,304 & 365 \\
 Armenian & 1,434 & 211 \\
\end{tabular}
\caption{Language statistics of our new cross-linguistic dataset on agreement relations. Most previous work has focused on English. ``\# cloze'' is the number of cloze examples curated for each language, and ``\# feature bundles'' is the number of unique sets of morphosyntactic features harvested for word types in BERT's vocabulary.}
\end{table}

\subsection{Feature data}

Our feature data comes from both the UD and the UniMorph projects \cite[downloaded June 2019]{sylak2016composition}. The UniMorph project also uses a consistent schema across all languages to annotate word types with morphological features. Although this schema is not the same as that used in UD, there is a deterministic mapping between the two \cite{mccarthy2018marrying}.

In this work, a word can take on a particular bundle of feature values (e.g.\ singular, feminine and third person) if it appears with those features in either UD or UniMorph. The UniMorph data directly specifies what bundles of feature values a word can take on. For the Universal Dependencies data, we say a word can take on a particular bundle if we ever see it with that bundle of feature values in a Universal Dependencies corpus for that language. Both sources individually allow for a word to have multiple feature bundles (e.g.\ \emph{sheep} in English can be singular or plural). In these cases, we keep all possible feature bundles. Finally, we filter out words that do not appear in BERT's vocabulary.


\section{Experiment}

Our experiment is designed to measure BERT's ability to model syntactic structure. Our experimental set up is an adaptation of that of \citet{goldberg2019assessing}. As in previous work, we mask one word involved in an agreement relation and ask BERT to predict it. \citet{goldberg2019assessing}, following \citet{linzen2016assessing}, considered a correct prediction to be one in which the masked word receives a higher probability than other inflected forms of the lemma. For example, when \emph{dogs} is masked, a correct response gives more probability to \emph{dogs} than \emph{dog}. This evaluation leaves open the possibility that selectional restrictions or frequency are responsible for the results rather than sensitivity to syntactic structure \cite{N18-1108}. To remove this possibility, we take into account all words of the same part-of-speech as the masked word. Concretely, we consider a correct prediction to be one in which the average probability of all possible correct words is higher than that of all incorrect words. By ``correct words'', we mean words with the exact same feature values and the same part of speech as the masked word. By ``incorrect words'', we mean words of the same part of speech as the masked word but that differ from the masked word with respect to at least one feature value. We ignore cloze examples in which there are fewer than 10 possible correct and 10 incorrect answers in our feature data. The average example in our cloze data is evaluated using 1,468 words, compared with 2 in \citet{goldberg2019assessing}.

Following \citet{goldberg2019assessing}, we use the pre-trained BERT models from the original authors\footnote{\url{https://github.com/google-research/bert}}, but through the PyTorch implementation.\footnote{\url{https://github.com/huggingface/pytorch-pretrained-BERT}} \citet{goldberg2019assessing} showed that in his experiments the base BERT model performed better than the larger model, so we restrict our attention to the base model. For English, we use the model trained only on English data, whereas for all other languages we use the multilingual model. 

\section{Results}

\begin{figure}[!ht]
    \centering
    \includegraphics[width=0.5\textwidth]{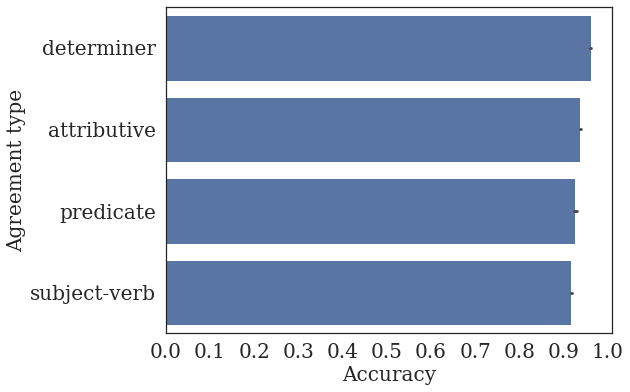}
    \caption{Accuracy per agreement type aggregated across all languages. In all four types, BERT performed above 90\% accuracy. Accuracy is slightly lower for predicate adjectives and subject-verb agreement relations, which typically have longer distance dependencies. Error bars are bootstrapped 95\% confidence intervals.}
    \label{figure:performance-by-type}
\end{figure}

Overall, BERT performs well on our experimental task, suggesting that it is able to model syntactic structure. BERT was correct in 94.3\% of all cloze examples. This high performance is found across all four types of agreement relations. Figure \ref{figure:performance-by-type} shows that BERT performed above 90\% accuracy in each type. Performance is best on determiner and attributive agreement relations, while worst on subject-verb and predicate adjective.

\begin{figure}[!ht]
    \centering
    \includegraphics[width=0.5\textwidth]{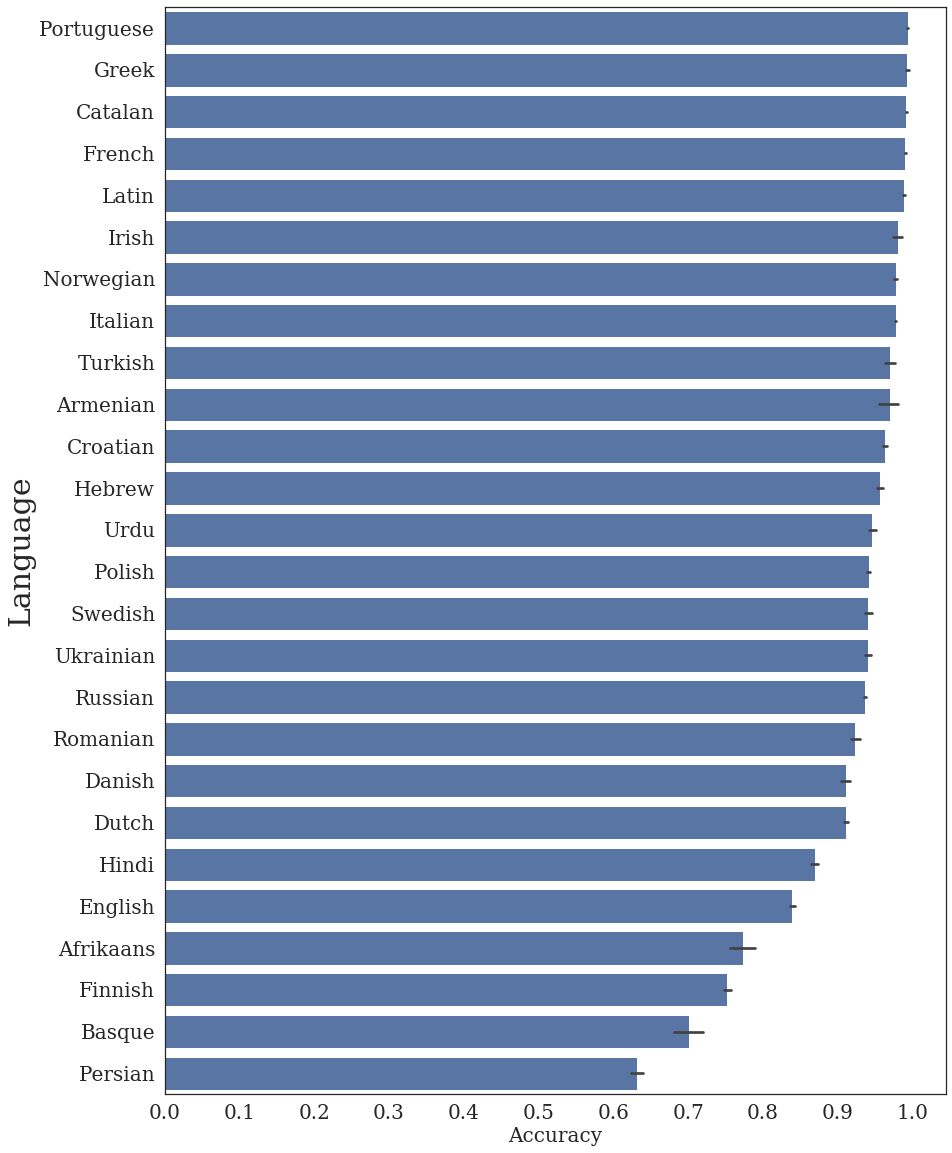}
    \caption{Accuracy per language aggregated across all four agreement types. In all \numLangs languages, BERT performs above 60\% accuracy. In most languages BERT performs above 90\% accuracy, although performance is significantly lower for a handful of languages. Error bars are bootstrapped 95\% confidence intervals.}
    \label{figure:performance-by-language}
\end{figure}

In figure \ref{figure:performance-by-language}, we see BERT's performance for each language. BERT performs well for the majority of languages, although some fare much worse than others. It is important to note that it is an unfair comparison because even though the datasets were curated using the same methodology, each language's dataset is different. It is possible, for example, that the examples we have for Basque are simply harder than they are for Portuguese.

\begin{figure}[!ht]
    \centering
    \includegraphics[width=0.5\textwidth]{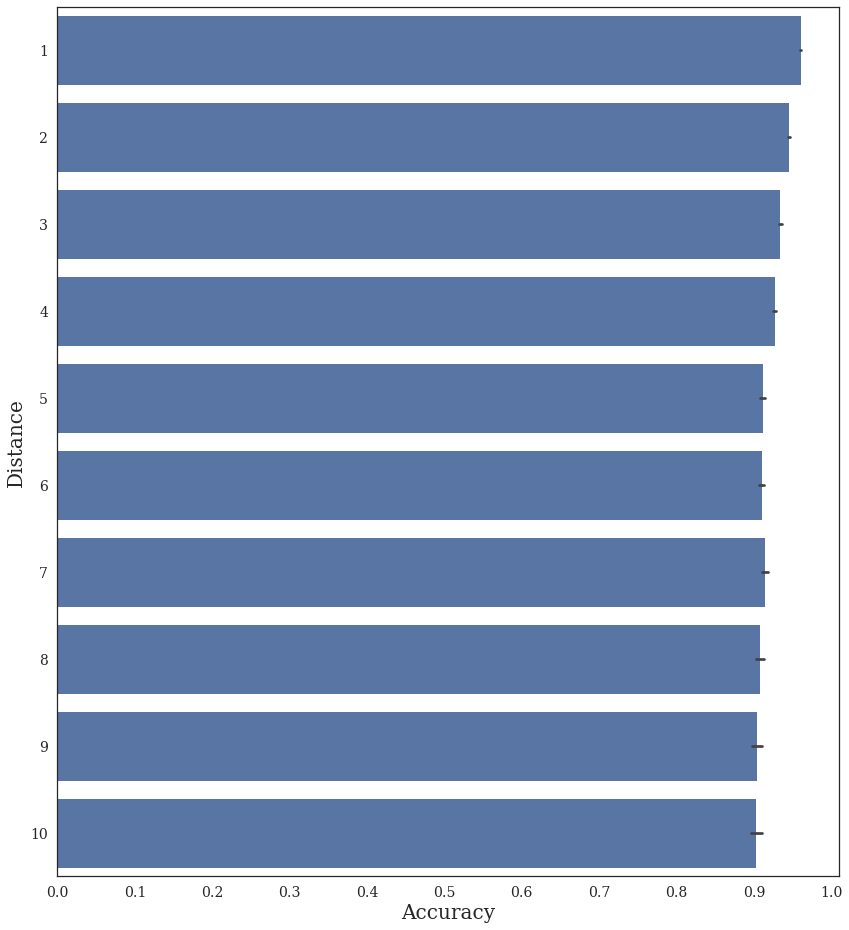}
    \caption{Accuracy as a function of distance between controller and target of agreement, aggregated across all languages and agreement types. BERT is relatively robust to longer-distance dependencies but does show a small decrease as the dependency length increases. Error bars are bootstrapped 95\% confidence intervals.}
    \label{figure:performance-by-distance}
\end{figure}

\begin{figure}[!htp]
    \centering
    \includegraphics[width=0.5\textwidth]{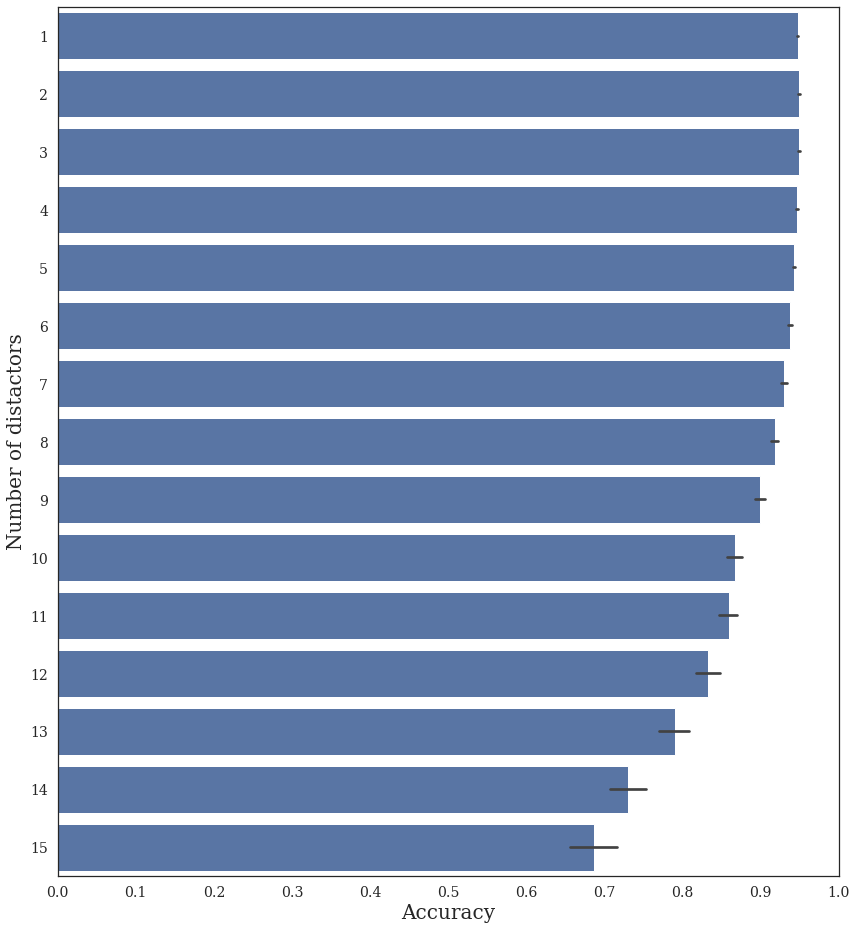}
    \caption{Accuracy as a function of number of distractors (other nouns in the sentence with different feature values), aggregated across all languages and agreement types. As with distance, BERT is quite robust to distractors although there is a more noticeable decrease in accuracy as more distractors are present. Error bars are bootstrapped 95\% confidence intervals.}
    \label{figure:performance-by-distractors}
\end{figure}

Finally, we ask how BERT's performance is affected by distance between the controller and the target, as well as the number of distractors. Figure \ref{figure:performance-by-distance} shows BERT's performance, aggregated over all languages and types, as a function of the distance involved in the agreement, while figure \ref{figure:performance-by-distractors} shows the same for number of distractors. There is a slight but consistent decrease in performance as the distance and the number of distractors increase. The decline in performance begins later in figure \ref{figure:performance-by-distractors} but drops more rapidly once it does.

\section{Related work}

Given the success of large pre-trained language representation models on downstream tasks, it is not surprising that that the field wants to understand the extent of their linguistic knowledge.\footnote{For a thorough overview of the recent push to understand what pre-trained models learn about language, see \citet{belinkov2019analysis}.} In our work, we looked exclusively at the predictions BERT makes at the word level. \citet{tenney2019bert} and \citet{jawahar2019does} examined the internal representations of BERT to find that syntactic concepts are learned at lower levels than semantic concepts. \citet{hewitt2019structural} are also interested in syntactic knowledge and propose a method to evaluate whether entire syntax trees are embedded in a linear transformation of a model's word representation space, finding that BERT does capture such information. As a complementary approach, \citet{clark2019does} studied the attention mechanism of BERT, finding clear correlates with interpretable linguistic structures such as direct objects, and suggest that BERT's success is due in part to its syntactic awareness. However, by subjecting it to existing psycholinguistic tasks,  \citet{ettinger2019bert} found that BERT fails in its ability to understand negation. In concurrent work, \citet{marten2019quantity} show that BERT does not consistently outperform LSTM-based models on English subject-verb agreement tasks.

\section{Conclusions \& future work}

Core linguistic phenomena depend on syntactic structure. Yet current state-of-the-art models in language representations, such as BERT, do not have explicit syntactic structural representations. Previous work by \citet{goldberg2019assessing} showed that BERT captures English subject-verb number agreement well despite this lack of explicit structural representation. We replicated this result using a different evaluation methodology that addresses shortcomings in the original methodology and expanded the study to \numLangs languages. Our study further broadened existing work by considering the most cross-linguistically common agreement types as well as the most common morphosyntactic features. The main result of this expansion into more languages, types and features is that BERT, without explicit syntactic structure, is still able to capture syntax-sensitive agreement patterns well. However, our analysis highlights an important qualification of this result. We showed that BERT's ability to model syntax-sensitive agreement relations decreases slightly as the dependency becomes longer range, and as the number of distractors increases. We release our new curated cross-linguistic datasets and code in the hope that it is useful to future research that may probe why this pattern appears.

The experimental setup we used has some known limitations. First, in certain languages some of the cloze examples we studied contain redundant information. Even when one word from an agreement relation is masked out, other cues remain in the sentence (e.g.\ when masking out the noun for a French attributive adjective agreement relation, number information is still available from the determiner). To counter this in future work, we plan to run our experiment twice, masking out the controller and then the target. Second, we used a different evaluation scheme than previous work \cite{goldberg2019assessing} by averaging BERT's predictions over many word types and plan to compare both schemes in future work.

\bibliography{emnlp-ijcnlp-2019}
\balance
\bibliographystyle{acl_natbib}


\end{document}